\documentclass{article}

\usepackage[preprint]{nips_2018}

\usepackage[utf8]{inputenc} 
\usepackage[T1]{fontenc}    
\usepackage{url}            
\usepackage{booktabs}       
\usepackage{amsfonts}       
\usepackage{nicefrac}       
\usepackage{microtype}      
\usepackage{amsmath}
\usepackage{xcolor}
\usepackage{enumitem}
\usepackage{tikz}
\newcommand{\reals}{\mathbb{R}}
\usepackage{caption}
\usepackage{subcaption}
\usepackage{wrapfig,lipsum,booktabs}
\usepackage{hyperref}
\usepackage{changepage}
\usepackage{comment}
\usepackage{titlesec}
\definecolor{mydarkblue}{rgb}{0,0.08,0.45}
\hypersetup{
        colorlinks=true,
        citecolor=mydarkblue,
        linkcolor=black,
    }

\title{Kernel-based Outlier Detection using the Inverse Christoffel Function}

\author{
  Armin Askari\thanks{Equal contribution}\\
  UC Berkeley\\
  \texttt{aaskari@berkeley.edu}
  \And
  Forest Yang\footnotemark[1]\\
  UC Berkeley\\
  \texttt{forestyang@berkeley.edu}
  \And 
  Laurent El Ghaoui \\
  UC Berkeley\\
  \texttt{elghaoui@berkeley.edu}
}
\begin{document}

\maketitle

\begin{abstract}
  Outlier detection methods have become increasingly relevant in recent years due to increased security concerns and because of its vast application to different fields. Recently, \citet{lasserre} noticed that the sublevel sets of the inverse Christoffel function accurately depict the shape of a cloud of data using a sum-of-squares polynomial and can be used to perform outlier detection. In this work, we propose a kernelized variant of the inverse Christoffel function that makes it computationally tractable for data sets with a large number of features. We compare our approach to current methods on 15 different data sets and achieve the best average area under the precision recall curve (AUPRC) score, the best average rank and the lowest root mean square deviation.
\end{abstract}

\section{Introduction}
With the rise of big data and the resurgence of machine learning, it has become increasingly important to be able to classify and categorize data based on its features. In spite of this, there is a large amount of data that does not follow regular observations. These data are considered \textit{anomalies} or \textit{outliers}. Learning how to discriminate outliers from regular observations in a data set can often translate into critical information in a variety of scenarios. For example, anomalous credit card bills could be a red flag for fraud or identity theft and anomalous patterns in sensor data for cyberphysical systems could mean that the system has been hacked \citep{Hodge2004}. In addition, outlier detection methods have been shown to make modelling tasks more accurate, making them invaluable for data cleaning \citep{Domingues}.\\

Outlier detection has been studied in statistics as early as the 1800s \citep{chandola}. Both generic and application-specific methods have been developed over time; a non-exhaustive list includes classification based, clustering based, nearest neighbor based, information theoretic, statistical, spectral, and distance based methods \citep{chandola}.  A recent paper by \citet{lasserre} proposes an outlier detection method based on the empirical inverse moment matrix and the inverse Christoffel function. Since the inverse Christoffel function is constructed based on empirical moments associated with the data points, \citet{lasserre} are able to provide theoretical properties and empirical results that show the efficacy of their method -- for more information on properties of the inverse Christoffel function, we direct the reader to \citet{lasserre} and references therein. In spite of this, the main computational bottleneck of their method comes from inverting the moment matrix whose size grows exponentially with the number features. For this reason, their method can only be applied to problems with a moderate number of features as opposed to data sets with a large number of features, which is typical of modern machine learning settings \citep{guyon}.
 
\paragraph{Outline and main contributions} In this paper, we extend the work of \citet{lasserre} by analyzing a lower bound on the inverse Christoffel function which is computationally tractable for outlier detection with a large number of features.  Our contribution is two-fold: 1) we show how our lower bound on the inverse Christoffel function can be interpreted in the context of classical $\ell_2$-regularized least squares and 2) how our modification can be kernelized. The kernelization achieves two important results: first, it eliminates the exponential dependence on the number of features, instead requiring a matrix inversion that has a cubic dependence on the number of data points. For data with several hundred features, the original method of directly inverting the moment matrix could not be completed due to memory shortage, whereas the kernelized method on the same data is able to run to completion. Second, it allows us to utilize other kernels for the inverse Christoffel function, notably kernels corresponding to infinite dimensional feature maps such as the radial basis function (RBF) kernel, which is impossible in the original formulation. The paper is structured as follows:

\begin{itemize}[leftmargin=*]
    \item \emph{Background}. Section \ref{sec:bkgd} outlines the notation and method used by \cite{lasserre}.
    \item \emph{Lower bound formualation and Kernelized Variant}. Section \ref{sec:analysis} introduces a lower bound on the method in Section 2 and how our method can be kernelized.
    \item \emph{Experiments}. Section \ref{sec:expt} introduces the proposed outlier detection method as well as traditional outlier detection methods, and the experimental setup involving 15 different data sets.
    \item \emph{Results}. Section \ref{sec:res} compares the performance of our method against other methods and the results are summarized in Table \ref{tab:auprcs}.
\end{itemize}

\section{Background}\label{sec:bkgd}


We are given a data set of $n$ points $x_i \in \reals^p$, $i=1,\ldots,n$ and we are interested in determining out of these $n$ points, which of them can we classify as outliers. Let $d$ denote a given natural number.
For $x \in \reals^p$, we define $v(x) \in \reals^s$ with $s$ is defined below: 
\begin{align}
s = \left(\begin{array}{c} p+d \\ d \end{array}\right)
\end{align}
to be a vector containing all the powers of the elements of $x$ up to and including degree $d$, arranged in lexicographical order, scaled in a specific way; namely
\vspace{0.5em}\begin{align}\label{eq:vx}
v(x) = (\sqrt{C(\alpha)}x^\alpha)_{|\alpha| \le d},
\end{align}
where $\alpha \in \mathbb{N}^p$ is an integer vector, $C(\alpha)$ is the corresponding multi-nomial coefficient, and $|\alpha| \le d$ refers to all integer vectors which sum up to $d$ or less -- simply put, $v(x)$ is the feature map of the degree $d$ polynomial kernel. With this definition of $v(x)$ we have
\begin{equation}\label{eq:vxvy}
	\forall x,y \in \reals^d ~:~ v(x)^Tv(y) = (1+x^Ty)^d.
\end{equation}
We denote by $D$ the $s \times s$ diagonal matrix containing the coefficients $\sqrt{C(\alpha)}$ on the diagonal, arranged in the same lexicographical order. Then, the moment matrix of the set of points $x_i \in \reals^p$, $i=1,\ldots,n$, as defined in \citet{lasserre}, can be expressed as $D^{-1}MD^{-1}$, where
\begin{align}
M := \frac{1}{n} \sum_{i=1}^n v(x_i)v(x_i)^T = VV^T,
\end{align}
with $V$ the $s \times n$ matrix
\begin{align}
V := \frac{1}{\sqrt{n}} \left( (v(x_1) , \ldots , v(x_n) \right).
\end{align}
Similarly, the \emph{inverse Christoffel function} in \citet{lasserre} can be expressed at any given point $x \in \reals^p$ as
\begin{align}
q(x) :&= \Big(D^{-1}v(x)\Big)^T\Big(D^{-1}MD^{-1}\Big)^{-1}\Big(D^{-1}v(x)\Big) \label{eq:useless}\\
&= v(x)^T M^{-1} v(x) \label{eq:cf}
\end{align}

where we assume that the moment matrix $M$ is positive-definite, which is the same as requiring that the set of points $x_i \in \reals^p$, $i=1,\ldots,n$ does not correspond to the zero set of some $d$-degree multivariate polynomial. For reasons that will become apparent in Section \ref{sec:analysis}, we will refer $q$ in \eqref{eq:cf} as the \textit{non-kernelized inverse Christoffel function}.

In \citet{lasserre}, the authors note that the level sets of \eqref{eq:cf} follows closely the shape of the cloud of points encoded in $x_i \in \reals^p$, $i=1,\ldots,n$; the mathematical justification for this can be found in the aforementioned paper and references therein. They use this function for outlier detection, labeling a point $x$ with large value of $q(x)$ as an outlier. In practice, since the matrix $M$ is large (its row and column size is $s$, which grows as $p^d$ for fixed $d$), the simple task of evaluating $q(x)$ for a given $x \in \reals^p$ becomes computationally intractable when $p$ is large, even for moderate values of $d$.

\section{Lower bound formulation and Kernelized Variant}\label{sec:analysis}

In order to make the method feasible for large scale problems, we consider a lower bound on \eqref{eq:cf} which converges to \eqref{eq:cf} and can be computed much more efficiently. We also show how our lower bound can be kernelized and conclude by showing how the method can be interpreted as a ridge regression problem. In Section \ref{sec:expt} we demonstrate how our lower bound compares to \eqref{eq:cf} as well as other traditional methods. 

\paragraph{A lower bound} 
\label{par:heuristic_based_on_monomials}
Let $x_0 \in \reals^p$ and $v := v(x_0) \in \reals^s$ as defined in \eqref{eq:vx}.  We seek to check the condition $q(x_0) \ge \delta$ (with $\delta >0$ given). We have that, 
\vspace{0.5em}\begin{align}
q(x_0) = v^T(VV^T)^{-1}v.
\end{align}
Instead of computing $q(x_0)$, we focus on computing a lower bound. We have for any $\rho >0$:
\vspace{0.5em}\begin{align} \label{eq:lb}
q(x_0) = v^T(VV^T)^{-1} v \ge v^T(\rho I+VV^T)^{-1}v = \rho^{-1} \phi(\rho),
\end{align}
where
\vspace{0.5em}\begin{align}
\phi(\rho) := v^T(I+ \rho^{-1}VV^T)^{-1} v .
\end{align}
The inequality means that if we seek to ascertain if a given point $x_0$ is an outlier, via the condition $q(x_0) \ge \delta$ (for $\delta>0$ given), then the condition 
\vspace{0.5em}\begin{align}
\phi(\rho) \ge \delta \rho
\end{align}
is sufficient for tagging $x_0$ as an outlier. The condition is equivalent in the limit $\rho \rightarrow 0$. To compute $\phi(\rho)$ efficiently, for a given $\rho>0$, we use the matrix inversion lemma:
\vspace{0.5em}\begin{align}
(I+\rho^{-1}VV^T)^{-1} = I-V(\rho I+ V^TV)^{-1}V^T,
\end{align}
which leads to the expression
\vspace{0.5em}\begin{equation}\label{eq:phi-kernel}
\phi(\rho) = \gamma - g^T(\rho I + G)^{-1}g,
\end{equation}
where
\vspace{0.5em}\begin{align}
G := V^TV \in \reals^{n \times n}, \;\; g:= V^Tv \in \reals^n, \;\; \gamma := v^Tv.
\end{align}
It follows from~(\ref{eq:vxvy}) that the $((i+1),(j+1))$ entry of the $(n+1)\times(n+1)$ matrix 
\[
\begin{bmatrix}
    \gamma       & g^T  \\
    g       & G 
\end{bmatrix}
\]
is given by
\vspace{0.5em}\begin{align}\label{eq:ker_fun}
(1+x_i^Tx_j)^d, \;\; 0 \le i,j \le n,
\end{align}
and similarly $\gamma=(1+x^Tx)^d$ and the $i$th entry of $g$ is $(1+x_i^Tx)^d$.
Thus, the triple $(G,g,\gamma)$ can be computed in $O(n^2 p\log d)$ time. In practice, $d$ is no larger than 5 so we can effectively drop the $\log d$ factor in the runtime. Hence the entire computation of $\phi(\rho)$ via the expression~(\ref{eq:phi-kernel}) grows as $O(n^3+n^2 p)$ whereas the method proposed by \citet{lasserre} takes $O(s^3)$ time, with the performance bottleneck being a required inversion of a dense $s\times s$ matrix.  

\paragraph{Using different kernels}
From \eqref{eq:phi-kernel}, we observe that all that is needed to calculate the inverse Christoffel function is the polynomial
kernel function $P:\mathbb{R}^p\times\mathbb{R}^p\to\mathbb{R},\, P(x,y)=(1+x^Ty)^d$. Note that the polynomial kernel can be substituted with any arbitrary kernel $K:\mathbb{R}^p\times\mathbb{R}^p\to \mathbb{R}$
; a natural choice is the RBF kernel, $R:\mathbb{R}^p\times
\mathbb{R}^p\to\mathbb{R},\, R(x,y)=\exp \{\-\|x-y\|^2/2\sigma^2\}$. The analogous inverse Christoffel function in \citet{lasserre} would be constructed by replacing the polynomial feature map $v:\mathbb{R}^p\to\mathbb{R}^s$ with the feature map $\psi$ associated to the kernel $K$. For an arbitrary kernel $K$ with feature map $\psi$, we have that \eqref{eq:phi-kernel} evaluated on the input $x$ is
\vspace{0.5em}\begin{align}\label{eq:q_kernel}
q_\psi(x) = \psi(x)^T\psi(x) - \psi(x)^T\Psi(\rho I+\Psi^T\Psi)^{-1}\Psi^T\psi(x), 
\end{align}

where the $i$-th column of $\Psi$ is $\psi(x_i)$ and where each of the inner products can be calculated following the methodology in  \eqref{eq:phi-kernel}-\eqref{eq:ker_fun} but replacing \eqref{eq:ker_fun} by the kernel $K$. By the same matrix inversion lemma, this is equal to  $(I+\rho^{-2}\Psi\Psi^T)^{-1}$. Note that in the case of infinite dimensional feature maps (for example when using the RBF kernel), directly computing the inverse Christoffel function is already impossible. We refer to $q$ in \eqref{eq:q_kernel} as the \textit{kerneliezd inverse Christoffel function}.

\paragraph{Interpretation as a subspace method} 
\label{par:interpretation_as_a_subspace_method}
We make the observation that our lower bound is in fact the optimal value of a ridge regression problem (see Appendix A)
\vspace{0.5em}\begin{align}\label{eq:ridge}
\phi(\rho) = v^T(I+\rho^{-1}VV^T)^{-1} v = \min_\theta \: \|V\theta - v\|_2^2 + \rho \|\theta\|_2^2 
\end{align}

In this sense, the use of the measure $\phi$ is natural; we tag a point as an outlier if it is far (in a regularized $l_2$-norm sense) from the subspace spanned by the columns of $V$. This reformulation of the problem allows us to calculate $\phi(\rho)$ by solving a ridge regression problem using iterative methods (such as the conjugate gradient method) which can be done much more efficiently for large amounts of data than computing the inverse of a dense matrix or solving a dense linear system. Furthermore, this interpretation as a subspace method naturally allows for robust variants where we assume the moment matrix $V$ is of the form $V = \tilde{V} + \Delta$ with $\Delta \in \mathcal{U}$ where $\mathcal{U}$ is an uncertainty set; \citet{chen2018} provide a structured way to construct this uncertainty set for robust outlier detection.

\section{Experiments} \label{sec:expt}
We compare the performance of the kernelized inverse Christoffel function as an unsupervised outlier detection method to the inverse Christoffel function proposed in \citet{lasserre} as well as to state of the art methods in \citet{sugiyama} and \citet{2016arXiv160500519L}. We show empirically on 15 data sets that our method achieves the best average performance and works in settings where computing the non-kernelized inverse Christoffel function is not possible.

\subsection{Methods}
The experiments follow the standard framework of outlier detection wherein a function $q: \mathbb{R}^p\to \mathbb{R}$
returns an outlierness score for each sample $x$ in the data set  $\mathcal{X}\subseteq \mathbb{R}^p$.
By choosing a threshold
$\delta$, we may obtain an outlier classifier $f_{q,\delta}:\mathbb{R}^p\to \{0,1\}$ as: 
\begin{align*}
    f_{q,\delta}(x) = \mathbf{1}\{q(x) \geq \delta\},
\end{align*}
where $\mathbf{1} \{ \cdot \}$ is the indicator function, and points mapped to 1 are labeled outliers. From here, precision-recall curves are generated by calculating the precision and recall of each classifier $f_{q,\delta}$ for $\delta\in \{q(x): x\in \mathcal{X}\}$. The Area Under the Precision-Recall Curve (AUPRC) is calculated from these classifiers and measures the effectiveness of an outlier detection method on a particular data set. AUPRC values are between 0 and 1, with higher values denoting a better classifier. \footnote{Experiments were done on a Lenovo Thinkpad X1Carbon with 8 GB of RAM and an Intel Core i7-5500U CPU @ 2.40GHz on Windows 8. Experiments were coded in Python 3.6 using standard packages, including numpy and sklearn \citep{scikit-learn}.} We outline below the different methods used in the experiments to calculate AUPRC values.

\paragraph{Kernelized Inverse Christoffel (KIC)} We set $q(x)$ as in \eqref{eq:q_kernel} and consider both the polynomial kernel and RBF kernel. For the polynomial kernel, we set the degree $d = 2$ and for the RBF kernel we set the lengthscale $\sigma = \sqrt{p}/2$; for both kernels we set $\rho = \|V^T V\|_F / C \sqrt{n}$ where $C = 500$. This is motivated by the fact that for large $\rho$ the shape of the data captured by the level sets of $q(x)$ is lost. We pick $\sigma = \sqrt{p}/2$ since for a normalized data set the average distance of a point from the origin is $\sqrt{p}$.

\paragraph{Kernelized Inverse Christoffel 2 (KIC2)} We construct $q(x)$ as in \eqref{eq:q_kernel} but only use a subset $\mathcal{X}'$ of the data $\mathcal{X}$ where $\mathcal{X}'$ consists of the points $x'$ such that 
\begin{align}
    |\{x\in \mathcal{X}|\,q(x) \leq q(x')\}| \leq \alpha n
\end{align} where $\alpha$ is taken to be 0.6. This can be thought of as a filtering step, motivated by the discussion in \citet{Hodge2004} of iteratively improving unsupervised outlier detection methods. We use $C=500$ for both kernels, $d=2$ for the polynomial kernel and $\sigma=\sqrt{p}/4$ for the RBF kernel.  

\paragraph{Inverse Christoffel (IC)} We set $q(x)$ as in \eqref{eq:cf}.

\paragraph{KNN} \citep{Ramaswamy:2000:EAM:335191.335437} devised the popular $k$-nearest neighbors technique which we use for outlier detection by setting $q(x) = d^k(x, \mathcal{X})$, where
$d^k(x, \mathcal{X})$ is the $k$th smallest value in $\{d(x,x'): x'\in \mathcal{X}\}$ and where $d$ is a metric. We use the standard value for $k=5$.

\paragraph{KSP} Proposed by \citep{sugiyama}, this method takes a random subset $S\subseteq \mathcal{X}$ with replacement and calculates the outlierness score of a point as the distance to the nearest neighbor in $S$. That is, $q(x) = d^1(x, S)$. As per the authors' suggestion, $|S|$ is taken to be 20.

\paragraph{KSP2} Like in KIC2, we perform a filtering step: we obtain an initial set of scores from KSP, let the $\alpha=0.5$ fraction of points with the lowest scores be $\mathcal{X}'$ and choose a new random subset $S'\subseteq \mathcal{X'}$ of size 20.  We then proceed with computing $q(x)$ as defined in KSP with $S'$. 

\paragraph{Influence (INF)}\hspace{-1mm} \citet{2016arXiv160500519L} assigns the outlierness score of a point to its influence, which is an upper bound on its sensitivity with respect to $k$-means clustering; we defer the reader to the paper for an exact description of their method. As stated in \citet{2016arXiv160500519L}, we average the influence scores over $k\in \cup_{i=1}^{15} \{500/i\}$ where $k$ is the number of cluster centers, with one slight caveat. To avoid errors, 500 is replaced with $\min(500, 0.8 n)$ where $n$ is the number of samples, because some data sets do not have enough samples to accommodate the previous set of $k$.

\newlength{\oldintextsep}
\setlength{\oldintextsep}{\intextsep}

\subsection{Data sets}
\setlength\intextsep{-1pt}
\begin{wraptable}{r}{6.5cm}
    \centering
    \begin{tabular}{lrrr}
\toprule
\textsc{Data set} &     $n$ &     $p$ &  \textsc{\#out} \\
\midrule
\textsc{Lymphography}            &   \textsc{148} &    \textsc{18} &           \textsc{6} \\
\textsc{Ionosphere}              &   \textsc{351} &    \textsc{34} &         \textsc{126} \\
\textsc{Mfeat1}                  &   \textsc{424} &   \textsc{649} &          \textsc{24} \\
\textsc{Arrhythmia}              &   \textsc{452} &   \textsc{274} &         \textsc{207} \\
\textsc{Mnist1}                  &   \textsc{524} &   \textsc{784} &          \textsc{24} \\
\textsc{WBC} &   \textsc{569} &    \textsc{30} &         \textsc{212} \\
\textsc{Mfeat2}                  &   \textsc{600} &   \textsc{649} &         \textsc{200} \\
\textsc{Mnist2}                  &   \textsc{750} &   \textsc{784} &         \textsc{250} \\
\textsc{Pima}                    &   \textsc{768} &     \textsc{8} &         \textsc{268} \\
\textsc{Isolet}                  &   \textsc{960} &   \textsc{617} &         \textsc{240} \\
\textsc{Gaussian}      &  \textsc{1000} &  \textsc{1000} &          \textsc{30} \\
\textsc{Letter}                  &  \textsc{1600} &    \textsc{32} &         \textsc{100} \\
\textsc{Optdigits}               &  \textsc{1688} &    \textsc{64} &         \textsc{554} \\
\textsc{Spambase}                &  \textsc{4601} &    \textsc{57} &        \textsc{1813} \\
\textsc{Annthyroid}              &  \textsc{7200} &     \textsc{6} &         \textsc{534} \\
\bottomrule
\end{tabular}
    \caption{Number of samples,  features, and number of outliers for each data set.}
    \label{tab:dsinfo}
\end{wraptable}
\setlength\intextsep{7pt}

We used a total of 15 data sets obtained from \citet{Dua:2017}, \citet{Rayana:2016}, and the official MNIST database. Each data set was normalized so that each feature had zero mean and unit variance. The data sets Optdigits, Isolet, Ionosphere, Wisconsin-Breast Cancer, Pima, and Spambase were prepared as in \citet{sugiyama} and \citet{2016arXiv160500519L}, deeming the samples in the smallest class as outliers and the rest as inliers. Arrhythmia was prepared as in \citet{sugiyama}, deeming the samples in the largest class as inliers and the rest as outliers. Mfeat1 (3, 9) and MNIST1 (4, 9) were prepared as in \citet{2016arXiv160500519L}, deeming the samples in the classes in parentheses as inliers and 3 samples from every other class as outliers. Mfeat2 (6, 9; 0) and MNIST2 (4, 9; 0) were prepared as in \citet{sugiyama}, using the samples belonging to the classes before the semicolon as inliers and the samples belonging to the class after the semicolon as outliers. For both MNIST1 and MNIST2, 250 samples from classes 4 and 9 were used, and for MNIST2, 250 samples from class 0 was used. The synthetic Gaussian data set was generated in a similar manner as in both works, generating 5 random Gaussian distributions and drawing 194 samples from each Gaussian to obtain the inliers. The Gaussians were generated by sampling means from $\mathcal{N}(0, I)$ and diagonal covariance matrices where each entry was sampled from $\mathcal{N}(0,1)$. The outliers were obtained by drawing 30 samples with each entry taken from the uniform distribution between the minimum and maximum value of the Gaussian samples drawn. The rest of the data sets, Lymphography, Letter, and Annthyroid, were taken from \citet{Rayana:2016} with outliers already specified.

\section{Results}\label{sec:res}
\begin{figure}[h!]
\begin{subfigure}{0.33\textwidth}
\centering
\includegraphics[trim={1.53cm 1.55cm 1.22cm 1.515cm},clip, width=0.8\linewidth]{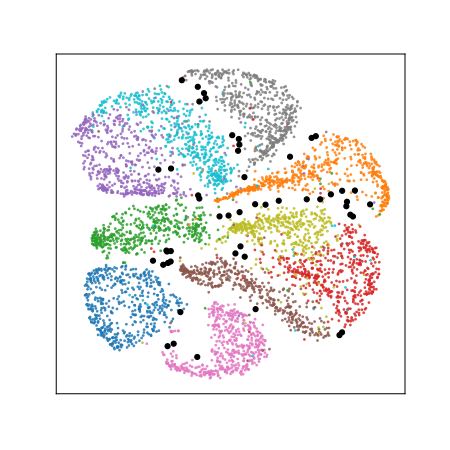}
\caption{KNN}
\end{subfigure}
\begin{subfigure}{0.33\textwidth}
    \centering
    \includegraphics[trim={1.53cm 1.55cm 1.22cm 1.515cm},clip, width=0.8\linewidth]{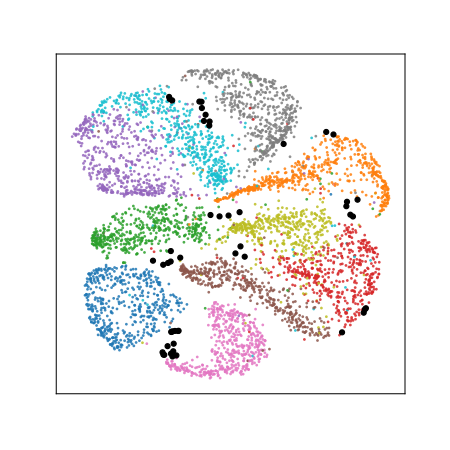}
    \caption{INF}
\end{subfigure}
\begin{subfigure}{0.33\textwidth}
\centering
\includegraphics[trim={1.53cm 1.55cm 1.22cm 1.515cm},clip, width=0.8\linewidth]{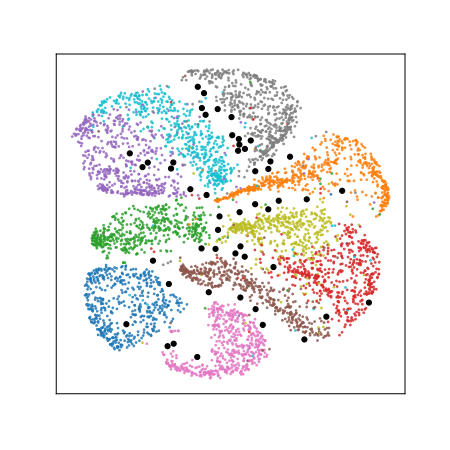}
\caption{KIC-RBF}
\end{subfigure} \\[1ex]
\begin{subfigure}{0.33\textwidth}
\centering
\includegraphics[trim={1.53cm 1.55cm 1.22cm 1.515cm},clip, width=0.8\linewidth]{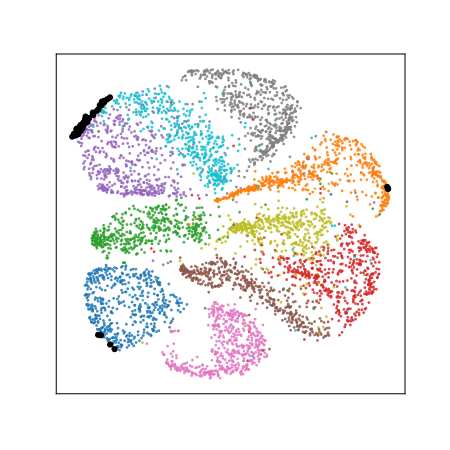}
\caption{IC, $d=2$}
\end{subfigure}
\begin{subfigure}{0.33\textwidth}
    \centering
    \includegraphics[trim={1.53cm 1.55cm 1.22cm 1.515cm},clip, width=0.8\linewidth]{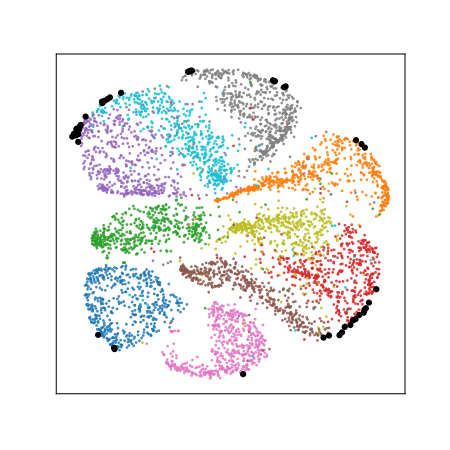}
    \caption{IC, $d=5$}
\end{subfigure}
\begin{subfigure}{0.33\textwidth}
    \centering
    \includegraphics[trim={1.53cm 1.55cm 1.22cm 1.515cm},clip, width=0.8\linewidth]{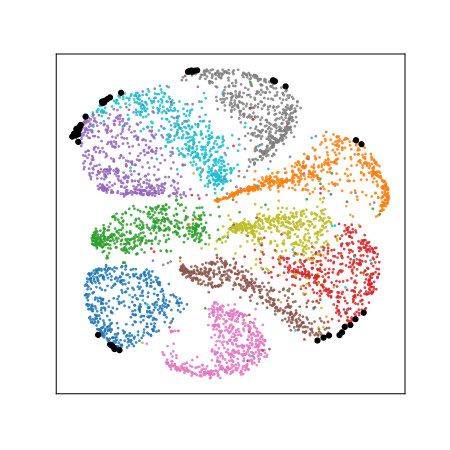}
    \caption{KIC, $d=5$}
\end{subfigure}
    
    \caption{The MNIST data set embedded in 2D using t-SNE. The color clouds correspond to digits (e.g. red cloud corresponds to digit one) and the black points are the 50 points with the highest outlierness scores according to the method underneath. For both KIC methods, we set $C=500$ and for KIC-RBF, we set $\sigma = 0.1$.}
\label{fig:tsne}
\end{figure}

\begin{figure}[h]
    \begin{subfigure}{0.33\textwidth}
\centering
\includegraphics[trim={1.53cm 1.55cm 1.22cm 1.515cm},clip, width=0.8\linewidth]{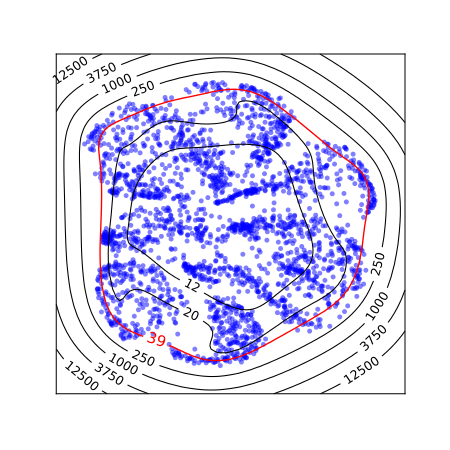}
\caption{IC, $d = 5$}
\end{subfigure}
    \begin{subfigure}{0.33\textwidth}
\centering
\includegraphics[trim={1.53cm 1.55cm 1.22cm 1.515cm},clip, width=0.8\linewidth]{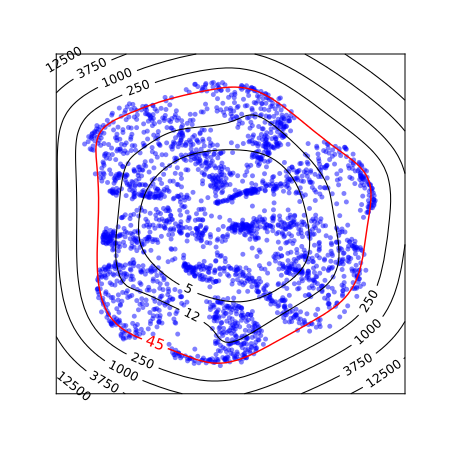}
\caption{KIC, $d = 5, C=500$}
\end{subfigure}
\begin{subfigure}{0.33\textwidth}
    \centering
    \includegraphics[trim={1.53cm 1.55cm 1.22cm 1.515cm},clip, width=0.8\linewidth]{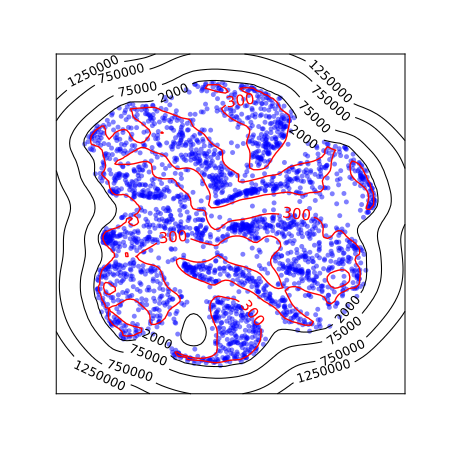}
    \caption{KIC-RBF, $\sigma = 1, C=500$}
\end{subfigure}
    \caption{Level sets of the inverse and kernelized inverse Christoffel function for the 2D projected MNIST data.}
    \label{fig:contours}
\end{figure}

Table \ref{tab:auprcs} contains the AUPRC values for each of the methods and data sets discussed in Section \ref{sec:expt}. Figure \ref{fig:tsne} shows the embedding of the MNIST data set in two dimensions via $t-$Distributed Stochastic Neighbor Embedding \citep{vandermaaten14a} in order to gain insight on how the outlier detection methods differ. Furthermore, to compare our kernelized method to the standard inverse Christoffel method, we plot the level sets of both outlierness score functions $q(x)$ in Figure \ref{fig:contours}. 

\begin{table}[t!]
\captionsetup{width=1.26\textwidth}
\begin{adjustwidth}{-2.5cm}{}
\begin{tabular}{lccccccccc}
\toprule
{} &   \textsc{KNN} &   \textsc{KSP} & \textsc{KSP2} & \textsc{KIC} &  \textsc{KIC2} &  \textsc{RBF} &  \textsc{RBF2} & \textsc{INF} &     \textsc{IC} \\
\midrule
\textsc{Lymphography}            &  \textsc{0.761} & \textsc{0.575$\pm$0.215}& \textsc{0.692$\pm$0.125} &    \textsc{0.710} &     \textsc{0.711} &        \textsc{0.548} &         \textsc{\textbf{0.766}} &  \textsc{0.715$\pm$0.063} &  \textsc{0.697}\\ 
\textsc{Ionosphere}              &  \textsc{0.936} & \textsc{0.835$\pm$0.049}&\textsc{0.834$\pm$0.058} &   \textsc{0.919} &     \textsc{0.920} &        \textsc{0.928} &         \textsc{0.932} &  \textsc{\textbf{0.959$\pm$0.008}} &  \textsc{0.918}\\
\textsc{Mfeat1}                  &  \textsc{0.724} &  \textsc{0.600$\pm$0.075} &\textsc{0.642$\pm$0.124}&    \textsc{0.685} &     \textsc{0.663} &        \textsc{0.713} &         \textsc{\textbf{0.741}} &  \textsc{0.678$\pm$0.034} &    \textsc{-}\\
\textsc{Arrhythmia}              &  \textsc{0.702} & \textsc{0.692$\pm$0.011}&\textsc{0.701$\pm$0.007} &    \textsc{0.704} &     \textsc{0.704} &        \textsc{0.699} &         \textsc{0.691} &  \textsc{\textbf{0.720$\pm$0.007}} &    \textsc{-}\\
\textsc{Mnist1}                  &  \textsc{\textbf{0.216}} &  \textsc{0.197
$\pm$0.019}&\textsc{0.196$\pm$0.009} &   \textsc{0.207} &     \textsc{0.202} &        \textsc{0.206} &         \textsc{0.214} &  \textsc{0.182$\pm$0.010} &    \textsc{-}  \\
\textsc{WBC}  &  \textsc{0.610}&  \textsc{0.600$\pm$0.068} &\textsc{\textbf{0.680$\pm$0.068}}&    \textsc{0.569} &     \textsc{0.594} &        \textsc{0.613} &         \textsc{0.618} &  \textsc{0.616$\pm$ 0.008} &  \textsc{0.676} \\
\textsc{Mfeat2}                  &  \textsc{0.221} &  \textsc{0.229$\pm$0.020} &  \textsc{\textbf{0.235$\pm$0.013}}&  \textsc{0.219} &     \textsc{0.215} &        \textsc{0.228} &         \textsc{0.220} &  \textsc{0.225$\pm$0.005} &    \textsc{-} \\
\textsc{Mnist2}                  &  \textsc{0.432} &  \textsc{0.457$\pm$0.058} &  \textsc{\textbf{0.487$\pm$0.060}}&  \textsc{0.439} &     \textsc{0.453} &        \textsc{0.419} &         \textsc{0.412} &  \textsc{0.425$\pm$0.007} &    \textsc{-}\\
\textsc{Pima}                    &  \textsc{0.530} &  \textsc{0.491$\pm$0.035} &\textsc{0.523$\pm$0.026}&    \textsc{0.493} &     \textsc{0.499} &        \textsc{0.524} &         \textsc{\textbf{0.547}} &  \textsc{0.537$\pm$0.006} & \textsc{0.493}\\
\textsc{Isolet}                  &  \textsc{0.408} & \textsc{0.401$\pm$0.118}&\textsc{0.483$\pm$0.178} &    \textsc{0.546} &     \textsc{\textbf{0.600}} &        \textsc{0.430} &         \textsc{0.433} &  \textsc{0.374$\pm$0.020} &    \textsc{-}\\
\textsc{Gaussian}      &  \textsc{\textbf{1.000}} & \textsc{0.983$\pm$0.029} & \textsc{0.999$\pm$0.005}&    \textsc{\textbf{1.000}} &     \textsc{\textbf{1.000}} &        \textsc{\textbf{1.000}} &         \textsc{\textbf{1.000}} &  \textsc{\textbf{1.000$\pm$0.000}} &    \textsc{-} \\
\textsc{Letter}                  &  \textsc{0.330} &  \textsc{0.136$\pm$0.026} & \textsc{0.121$\pm$0.014}&   \textsc{0.349} &     \textsc{0.280} &        \textsc{\textbf{0.383}} &         \textsc{0.353} &  \textsc{0.294$\pm$0.024} &  \textsc{0.355} \\
\textsc{Optdigits}               &  \textsc{0.208} &  \textsc{0.223$\pm$0.024} & \textsc{0.205$\pm$0.008}&   \textsc{0.226} &     \textsc{0.217} &        \textsc{0.216} &         \textsc{0.211} &  \textsc{0.212$\pm$ 0.003} &  \textsc{0.222}\\
\textsc{Spambase}                &  \textsc{0.402} &  \textsc{0.415$\pm$0.020} & \textsc{0.412$\pm$0.009}&  \textsc{0.410} &     \textsc{0.410} &        \textsc{0.372} &         \textsc{0.353} &  \textsc{\textbf{0.420$\pm$0.003}} &  \textsc{0.385}\\
\textsc{Annthyroid}              &  \textsc{0.228} &  \textsc{0.203$\pm$0.031} & \textsc{0.203$\pm$0.014}&   \textsc{0.191} &     \textsc{0.355} &        \textsc{0.230} &         \textsc{\textbf{0.267}} &  \textsc{0.200$\pm$0.004} &  \textsc{0.193}\\
\toprule
\textsc{Average}                 &  \textsc{0.514} &  \textsc{0.470} & \textsc{0.494}&   \textsc{0.511} &    \textsc{\textbf{0.522}} &        \textsc{0.501} &         \textsc{0.517} &  \textsc{0.504} &  \textsc{0.467} \\
\textsc{Avg. Rank}&\textsc{4.333}&\textsc{6.000}&\textsc{5.267}&\textsc{4.867}&\textsc{4.6}&\textsc{4.667}&\textsc{
\textbf{4.000}}&\textsc{4.2}&\textsc{5.375}\\
\textsc{RMSD} &\textsc{0.066}&\textsc{0.117}&\textsc{0.096}
&\textsc{0.062}&\textbf{\textsc{0.047}}&\textsc{0.084}&\textsc{0.059}&\textsc{0.081}&\textsc{0.069}\\
\bottomrule
\end{tabular}\\\\
\end{adjustwidth}
\caption{Performance measured by AUPRC on different data sets. RBF and RBF2 are shorthand for KIC-RBF and KIC-RBF2 respectively. \textsc{KIC} and \textsc{IC} both used degree $d = 2$ polynomial kernel. For the methods involving randomness, the AUPRC is averaged over 30 trials with the uncertainty being the standard deviation of the trials. The data sets for which the IC method ran into a MemoryError are denoted by a dash. The statistics at the bottom for IC are only considering data sets for which IC did not run into a MemoryError. Average rank is the rank from 1 to 9 (1 to 8 for data sets where IC errors) that each method obtains for each data set averaged together. RMSD is the root mean squared deviation and is calculated for each method by finding the square root of the average of the squared difference between the AUPRC obtained by the method and the best AUPRC obtained on the data set for each data set.}
\label{tab:auprcs}
\end{table}

The two dimensional embedding of MNIST consists of 40000 data points randomly sampled from the original data set of 60000 points. The contours in Figure \ref{fig:contours} were constructed using 2500 points sampled randomly from the 40000 used to generate Figure \ref{fig:tsne}. Note the data set was used for purely illustrative purposes and was not normalized, hence the non-standard values of $\sigma$. Furthermore we selected $\sigma = 1$ Figure \ref{fig:contours} as opposed to $\sigma = 0.1$ as in Figure \ref{fig:tsne} for visual clarity.

From Figure \ref{fig:contours} it is clear that KIC-RBF produces contours that fit the shape of the data much more closely than IC using a polynomial feature mapping. The red level sets are roughly those past which points are classified as outliers. The level sets of the inverse Christoffel methods that used the polynomial feature map did not closely match the shape of the 2D MNIST clusters, showing some distortion due to the data but still forming singular loops containing regions of white space. On the other hand, the RBF-based KIC method level set shown in red is able to trace the complicated shape of the data, accommodating disconnected clusters and rejecting white space.

This suggests the generality of the data fitting property to multiple kernels and may lead to a more unifying interpretation of why the inverse Christoffel function captures the shape of a cloud of points. In the MNIST t--SNE projected 2D example in Figure \ref{fig:tsne}, it is clear that the RBF kernel has greater ability to fit complicated but consistent patterns in data than the polynomial kernel. Even with a degree as high as 5, the polynomial kernel returns outliers than are on the outskirts of the data, failing to capture the variation within the data.  Our kernel formulation allows the inverse Christoffel function to capture complex data by giving it access to the RBF kernel, which is inaccessible to the original formulation due to the mapping corresponding to the RBF kernel being infinite dimensional. 

It is also worth noting that in Figure \ref{fig:contours} that the KIC contours are very similar to the IC contours. The corresponding contours for $C=5000$ are not shown but match the IC contours essentially exactly. For $C=500$, we observe contours that are similarly shaped to those of IC, but more circular. Furthermore, they are pushed slightly outwards, which we expected from \eqref{eq:lb}. This seems to indicate that our formulation does not deviate too much from that in \citet{lasserre} and yet is more computationally efficient and tractable for problems where IC is not.

\paragraph{Discussion} Table \ref{tab:auprcs} highlights the potential of KIC/KIC-RBF as a viable method for unsupervised outlier detection. As stated in \citet{lasserre}, the IC method cannot be applied to data sets with a moderate number of features due to its exponential dependence on the number of features. In fact, the data set with the largest number of features that IC was able to run on was Optdigits which had only 64 features. For the data sets with an order of magnitude larger features (such as MNIST with 784 features), with $d=2$, the polynomial feature map would map a 784-dimensional vector to a ${}_{786}C_2 = $308505-dimensional vector. The computation of $(VV^T)^{-1}$ would then require inverting a 308505 $\times$ 308505 dense matrix. This limits the power of the method proposed by \citet{lasserre}; for example on the synthetic Gaussian data set, every method was able to achieve a near perfect AUPRC whereas IC could not run. Furthermore, the KIC method was able to achieve higher AUPRCs on the same data sets than the IC method. This suggests that the $\rho I$ perturbation in \eqref{eq:lb} may play a beneficial role in outlier detection by serving as a regularizer as highlighted in \eqref{eq:ridge}.  

KIC based methods outperform the others on almost half of the data sets. The KIC methods perform best on the cumulative metrics: average AUPRC, average rank, and RMSD when compared to other state of the art methods. KIC2 achieved the highest average AUPRC out of the methods shown, as well as the lowest RMSD. Furthermore, KIC2-RBF achieved the best average rank.  Only the filtered versions of the KSP and KIC methods (KSP2, KIC2, RBF2) are shown in the table; the reason for this is because we observed a decline in performance upon applying the filtering step to KNN (by using the distances to the $k$th nearest filtered point) and INF (by sampling the cluster centers from the filtered points). 

It is also worth noting that the choices of $C=500$ and $\sigma=\sqrt{p}/2$ for KIC-RBF and $\sigma=\sqrt{p}/4$ for KIC2-RBF could be improved significantly for any particular data set. While the KIC-RBF methods do not perform the best in the above table, when one is allowed to tune the value of $\rho$ and $\sigma$ for each particular data set, they perform significantly better. With some tuning for each dataset, KIC2-RBF was able to achieve an average AUPRC greater than 0.65, which would dominate average scores in Table \ref{tab:auprcs}. Taking an extreme choice of $C=10, \sigma=\sqrt{p}/20$ for KIC-RBF and $\sigma = \sqrt{p}/25$ for KIC2-RBF, KIC2-RBF achieved AUPRCs of $0.979, 0.963$, and $0.722$ on Mfeat2, Isolet, and Optdigits respectively; the highest AUPRCs for these datasets in Table \ref{tab:auprcs} being $0.235, 0.600$, and $0.223$ respectively. This hints at the potential of the RBF kernel to fit to certain types of data sets especially well. 

\section{Conclusion}
In this paper we consider a modification of the inverse Christoffel function proposed by \citet{lasserre} to be used for outlier detection. Our method can be seen as a lower bound on the Christoffel function which we show makes the problem of outlier detection computationally feasible. Furthermore, we illustrate how our method is kernelizable, and by utilizing different kernel functions we are able to achieve state of the art AUPRC scores on traditional outlier data sets. In particular, the RBF kernel applied to the kernelized inverse Christoffel function shows potential for excelling at outlier detection on datasets which others do not do well on.

Possible extensions of this work include looking at randomized variants where the moment matrix is constructed from a subset of the data, using different regularizers when solving \eqref{eq:ridge}, considering robust variants, and looking at more structured ways of selecting hyper-parameters based on characteristics of different data sets. Since the best method for each dataset is quite varied in Table \ref{tab:auprcs}, future work may also attempt to characterize the types of datasets different methods do better on.

\newpage

\small
\bibliographystyle{apalike}
\bibliography{outlier}{}

\clearpage

\appendix
\section*{Appendix A  Optimal value as a ridge regression problem}\label{appendix}
In this section we prove (17). Since the problem is convex and unconstrained, the optimal solution results by simply taking the gradient of the objective with respect to $\theta$ and setting it to $0$. We find that

\begin{align}
    &V^T(V\theta^* - v) + \rho \theta^* = 0 \\
    \Rightarrow \;\;\; &\theta^\star = (V^T V + \rho I)^{-1} V^T v
\end{align}

Plugging back into the objective we have that
\begin{align}
    \|V \theta^* - v\|_2^2 + \rho \|\theta^*\|_2^2 &= (V\theta^* - v)^T (V\theta^* - v) + \rho \theta^T \theta   \\
    &= \theta^{*^T} \Big( V^T (V\theta^* - v) + \rho \theta^*\Big) + v^T v - \theta^{*^T} V^T v \\
    &= v^T \Big( I - V(V^T V + \rho I)^{-1} V^T \Big) v \\
    &= v^T ( I + \rho^{-1} VV^T ) v
\end{align}
where the third equality holds because of the optimality condition and the last equality by the matrix inversion lemma.

\end{document}